\documentclass[letterpaper]{article} 
\usepackage{aaai2026}  
\usepackage{times}  
\usepackage{helvet}  
\usepackage{courier}  
\usepackage[hyphens]{url}  
\usepackage{graphicx} 
\usepackage{natbib}  
\usepackage{caption} 
\usepackage{algorithm}
\usepackage{algorithmic}
\usepackage{multirow}
\usepackage{algorithm}
\usepackage{algorithmic}
\usepackage{booktabs, multirow, xcolor, colortbl}
\definecolor{mygray}{gray}{.92}
\definecolor{myblue}{RGB}{243,248,254}
\definecolor{mygreen}{RGB}{245,248,238}
\definecolor{myred}{RGB}{245, 231, 234}
\newcommand{\bftab}{\fontseries{b}\selectfont}
\usepackage[table]{xcolor}
\usepackage{tikz}
\usepackage{adjustbox}
\usepackage{aaai2026}

\usepackage{newfloat}
\usepackage{listings}
\DeclareCaptionStyle{ruled}{labelfont=normalfont,labelsep=colon,strut=off} 
\floatstyle{ruled}
\newfloat{listing}{tb}{lst}{}
\floatname{listing}{Listing}
\usepackage{amssymb}
\usepackage{pifont}
\usepackage{graphicx}
\usepackage{amsmath}
\usepackage{bm}
\usepackage{amsthm}
\usepackage{mathrsfs}
\usepackage{array}
\usepackage{booktabs}

\title{MacVQA: Adaptive Memory Allocation and Global Noise Filtering for \\ Continual Visual Question Answering}

\author {
    Zhifei Li\textsuperscript{\rm 1,\rm 4,\rm 5},
    Yiran Wang\textsuperscript{\rm 1},
    Chenyi Xiong\textsuperscript{\rm 1}, 
    Yujing Xia\textsuperscript{\rm 1},
    Xiaoju Hou\textsuperscript{\rm 2,}\thanks{Corresponding Authors.}, \\
    Yue Zhao\textsuperscript{\rm 3,}, 
    Miao Zhang\textsuperscript{\rm 1,}\footnotemark[1],  
    Kui Xiao\textsuperscript{\rm 1,}\footnotemark[1], 
    Bing Yang\textsuperscript{\rm 1}
}
\affiliations {
    \textsuperscript{\rm 1}School of Computer Science, Hubei University, Wuhan 430062, China\\
    \textsuperscript{\rm 2}Institute of Vocational Education, Guangdong Industry Polytechnic University, Guangzhou 510300, China\\
    \textsuperscript{\rm 3}Shandong Police College, Ji’nan 250200, China\\
    \textsuperscript{\rm 4}Hubei Key Laboratory of Big Data Intelligent Analysis and Application (Hubei University), Wuhan 430062, China\\
    \textsuperscript{\rm 5}Key Laboratory of Intelligent Sensing System and Security (Hubei University), Ministry of Education, Wuhan 430062, China\\
    \{zhifei1993, zhangmiao, xiaokui\}@hubu.edu.cn, 2023030010@gdip.edu.cn,\\
    zhaoy@sdpc.edu.cn, \{yiranwang, xiongchenyi, xiayujing\}@stu.hubu.edu.cn, yangbing@126.com

}

\usepackage{bibentry}

\begin{document}

\maketitle

\begin{abstract}
Visual Question Answering (VQA) requires models to reason over multimodal information, combining visual and textual data. With the development of continual learning, significant progress has been made in retaining knowledge and adapting to new information in the VQA domain. However, current methods often struggle with balancing knowledge retention, adaptation, and robust feature representation. To address these challenges, we propose a novel framework with adaptive memory allocation and global noise filtering called MacVQA for visual question answering. MacVQA fuses visual and question information while filtering noise to ensure robust representations, and employs prototype-based memory allocation to optimize feature quality and memory usage. These designs enable MacVQA to balance knowledge acquisition, retention, and compositional generalization in continual VQA learning. Experiments on ten continual VQA tasks show that MacVQA outperforms existing baselines, achieving 43.38\% average accuracy and 2.32\% average forgetting on standard tasks, and 42.53\% average accuracy and 3.60\% average forgetting on novel composition tasks.
\end{abstract}

\begin{links}
    \link{Code}{https://github.com/HubuKG/MacVQA}
\end{links}

\section{Introduction}
Visual Question Answering (VQA) integrates visual and textual information to answer questions about images. As a representative task in multimodal learning, VQA has garnered significant attention due to its wide range of real-world applications, including autonomous driving \cite{2}, medical diagnosis \cite{1}, accessibility tools for visually impaired individuals \cite{4}, and intelligent human-computer interaction systems \cite{3}. For instance, a VQA system can assist medical professionals in interpreting complex diagnostic imaging by answering questions about abnormalities or help autonomous vehicles analyze scenes through visual queries. These applications highlight the importance of VQA as a bridge between visual perception and natural question understanding.

\begin{figure}[t]
\centering
\includegraphics[width=1.0\columnwidth]{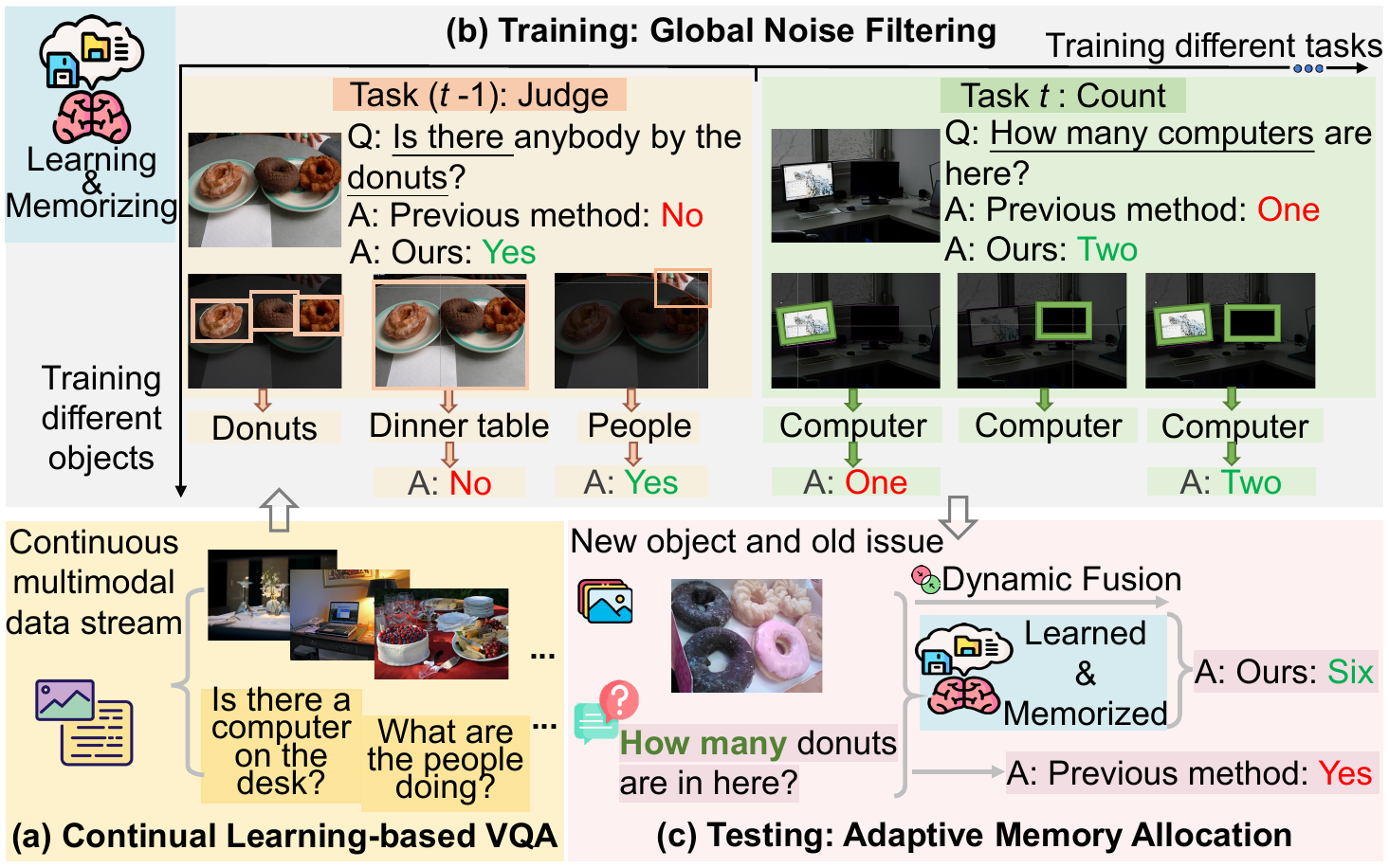} 
\caption{Illustration with continual learning methods for VQA. (a) Continual Learning-based VQA: The system learns new tasks while retaining prior knowledge. (b) Global Noise Filtering: Refines visual features by removing noise for better task performance. (c) Adaptive Memory Allocation: Combines learned and memorized knowledge for question answering.}
\label{figure:1}
\end{figure}

Continual learning (CL) has emerged as a promising paradigm in the VQA domain, offering solutions to the challenges posed by evolving environments and dynamic tasks \cite{5}. By enabling models to learn and adapt to new tasks without forgetting previously acquired knowledge \cite{6,7}, CL addresses the inherent requirements of VQA systems for long-term adaptability and scalability \cite{46,47}. Its value is particularly evident in scenarios where VQA systems need to handle diverse datasets, adapt to new visual concepts, and retain previously learned reasoning capabilities \cite{8,9}. The integration of CL into VQA not only enhances the robustness and flexibility of these systems but also paves the way for more practical and efficient applications in real-world settings.

Existing methods primarily adopt regularization-based or rehearsal-based methods \cite{10,50,51}. Regularization techniques preserve prior knowledge by protecting critical parameters, with Memory Aware Synapses (MAS) \cite{11} maintaining important weights and Neuronal Plasticity Control (NPC) \cite{12} regulating synaptic adaptability. Complementary rehearsal strategies reinforce learning through stored task memories, exemplified by Experience Replay (ER) \cite{13} and its enhanced variant Dark Experience Replay (DER) \cite{14}. Recent VQA-specific innovations include: VQACL \cite{15} evaluating unseen skill-concept combinations; PROOF \cite{18} employing expandable projections; QUAD \cite{16} eliminating visual storage via attention consistency; and ProtoGroup \cite{17} stabilizing representations through multi-prototype grouping with dual-score memory selection.

Despite these advancements, existing methods face significant limitations in multimodal continual learning. Regularization struggles with fine-grained interactions, while rehearsal depends heavily on memory replay, limiting scalability \cite{43,44,45,49}. VQACL and PROOF adapt poorly to dynamic tasks; QUAD is less robust to noisy features; ProtoGroup incurs computational overhead and is sensitive to clustering parameters. Such constraints hinder balancing retention, adaptation, and robust reasoning.

To address these challenges, we propose MacVQA, an innovative framework designed to enhance multimodal feature representation and optimize memory allocation for continual visual question answering, as illustrated in Figure~\ref{figure:1}. MacVQA integrates two key components: Global Noise Filtering (GonF) and Adaptive Memory Allocation (AMA). GonF improves multimodal feature quality by removing irrelevant noise and fusing global visual and linguistic information, ensuring robust feature representations for complex reasoning tasks. AMA employs adaptive memory strategies, including prototype retrieval and memory decay, to efficiently update knowledge, mitigate forgetting, and maintain scalability. Together, these components enable MacVQA to balance multimodal information adaptively while effectively integrating new knowledge and preserving prior learning.

In this paper, we have made the following contributions:

\begin{itemize}

\item We propose Global Noise Filtering, which combines global feature fusion and noise removal to enhance multimodal feature robustness and reasoning quality significantly.

\item We design Adaptive Memory Allocation, which uses prototype retrieval, memory decay, and adaptive updates to optimize knowledge allocation in dynamic continual learning environments.

\item We conduct experiments on ten tasks in VQA v2 dataset \cite{19}. MacVQA achieves accuracies of 43.38\% and 42.53\% on standard and novel task combinations, while reducing forgetting to 2.32\% and 3.60\%.

\end{itemize}

\section{Related Work}
\label{sec:2}
\subsection{Visual Question Answering}
Visual Question Answering (VQA) is a multimodal task that combines visual and textual inputs to answer questions about images. Early methods established visual-textual alignment through attention mechanisms \cite{22,23}, while transformer architectures \cite{24,25,26,27} advanced cross-modal fusion via self-attention. Recent generative paradigms unify diverse tasks under a text generation framework, eliminating specialized modules while enabling flexible, open-ended responses. This evolution reflects progressive architectural simplification from component-specific designs to unified processing.

However, as VQA applications expand to more complex and dynamic scenarios, traditional methods face challenges in handling diverse and evolving tasks \cite{39}. Multimodal data in real-world environments often exhibit high variability, requiring models to establish accurate connections between rich visual content and diverse linguistic expressions \cite{27,40}. For instance, many datasets highlight the difficulties posed by incomplete or noisy images \cite{41}, while open-ended tasks reveal limitations in predefined structures and fixed setups. Addressing these challenges requires models capable of adapting to dynamic data distributions and generalizing to unseen tasks \cite{28,38}, marking a crucial direction for future research in VQA.

\begin{figure*}[t]
\centering
\includegraphics[width=0.9\linewidth]{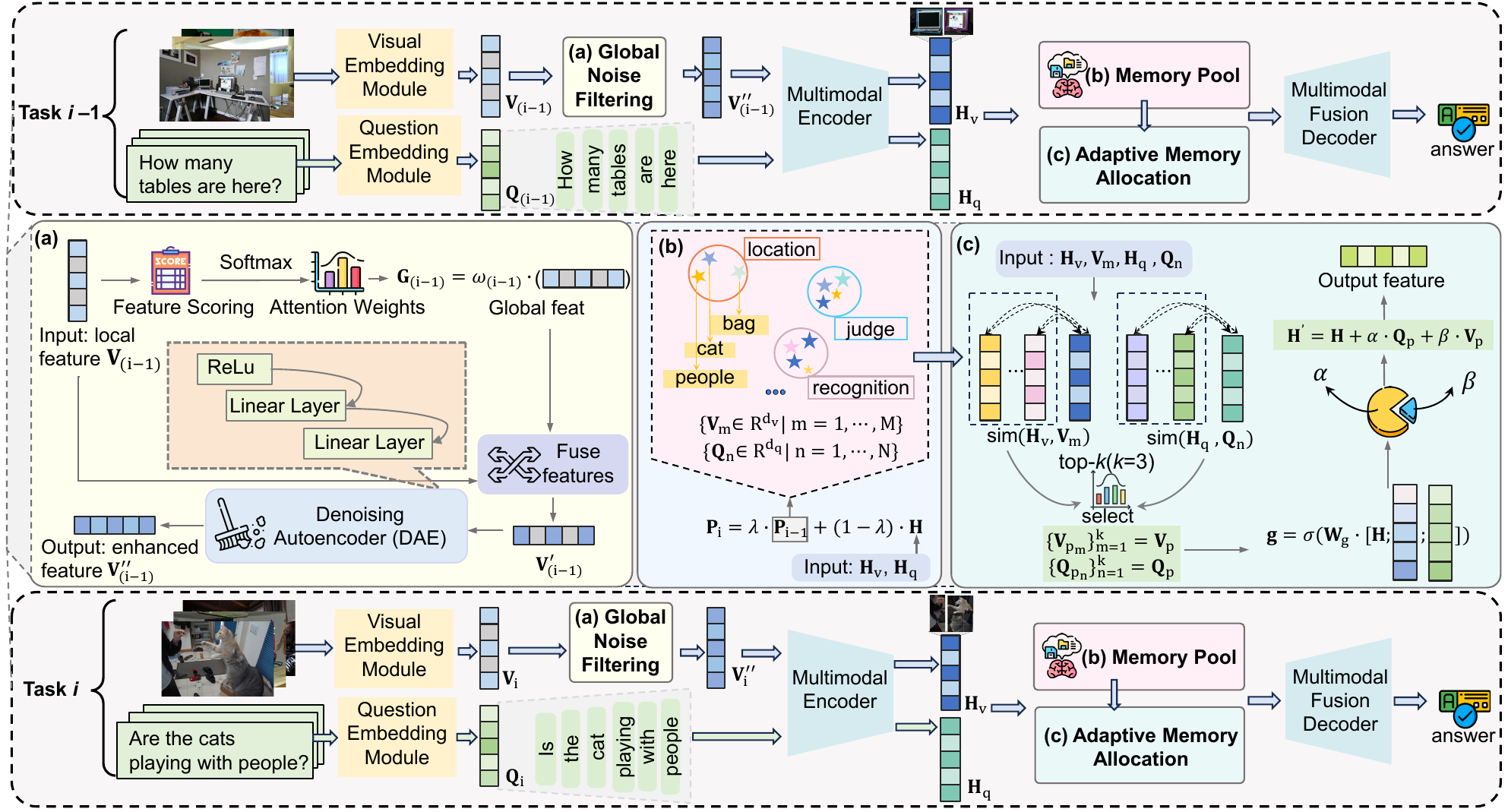} 
\caption{The proposed framework consists of three key modules: (a) Global Noise Filtering, which refines visual features by removing noise through scoring, fusion, and a Denoising Autoencoder (DAE); (b) Memory Pool, storing visual and textual prototypes as dynamic references; (c) Adaptive Memory Allocation, which dynamically allocates prototype features from the memory pool to enhance adaptability and generalization.}
\label{figure:2}
\end{figure*}

\subsection{Continual Learning in VQA}
Continual learning enables sequential knowledge acquisition while preserving prior representations. Current VQA continual learning methods primarily employ two paradigms. Regularization-based methods constrain parameter updates to protect critical knowledge, exemplified by MAS \cite{11} computing parameter importance via Fisher-based metrics, and NPC \cite{12} modulating layer-specific learning rates through activation stability. However, these methods struggle with modality conflicts in multimodal architectures. Rehearsal-based methods mitigate forgetting through stored exemplars, ER \cite{13} replays random past data (high storage), while DER \cite{14} augments ER via logit distillation (privacy-constrained).

Recent frameworks tailored for continual learning in VQA address these challenges. VQACL \cite{15} introduces a dual-level task sequence and compositional testing protocol; QUAD \cite{16} employs question-only replay and attention consistency distillation; while ProtoGroup \cite{17} leverages multi-prototype grouping for stable representations and dual-score memory selection. Nonetheless, challenges such as limited feature robustness, catastrophic forgetting, and computational overhead persist in dynamic environments. 

To fundamentally address these issues, our method enhances representation robustness and strengthens knowledge retention across the learning process. It enables stable task adaptation, preserves long-term knowledge, and supports generalization, offering a comprehensive solution for continual learning in dynamic visual question scenarios.

\section{Methodology}
\label{sec:3}
We propose a novel multimodal continual learning framework (Figure~\ref{figure:2}) for visual question answering, addressing feature representation, memory, and knowledge update challenges. By incorporating Global Noise Filtering and Adaptive Memory Allocation modules, our framework enhances feature robustness, memory efficiency, and task adaptation. Module details are described below.
\subsection{Global Noise Filtering}  
\label{sec:3.1}

The Global Noise Filtering (GonF) module addresses the problem of noisy and irrelevant region features in visual representation learning. By globally highlighting important information and filtering out noise, GonF improves the robustness and quality of visual features for visual-question tasks.

For each outer task (e.g., Count), the same process is applied to all its image-question pairs (inner tasks). Here, we take the $(i-1)$-th outer task as an example to illustrate the feature extraction and aggregation process.

Visual feature extraction employs a pre-trained Faster R-CNN \cite{31} model to extract $n$ region features from each image, represented as $\mathbf{V}_{(i-1),j} \in \mathbb{R}^{n \times d}$, where $n$ denotes the number of detected image regions, $d$ represents the feature embedding dimension (typically 2048), $(i-1)$ indicates the index of the previous (outer) task, and $j$ corresponds to the attention head index in multi-head architectures. Textual features are obtained by encoding input queries into embeddings $\mathbf{Q}_{(i-1),j} \in \mathbb{R}^{L \times d}$, with $L$ signifying the number of words in the query.

The noise filtering process begins with feature scoring, which assigns attention weights $\omega_{(i-1),j,m}$ to each image region through a linear transformation and softmax normalization:
\begin{equation}
\omega_{(i-1),j,m} = \frac{\exp(\mathrm{score}(\mathbf{V}_{(i-1),j,m}))}{\sum{l=1}^{n} \exp(\mathrm{score}(\mathbf{V}_{(i-1),j,l}))},
\label{eq:softmax}
\end{equation}
where $\omega_{(i-1),j,m}$ quantifies the attention weight for the $m$-th region in the $(i-1)$-th task and head $j$. These weights are then used to generate a global feature vector $\mathbf{G}_{(i-1),j} \in \mathbb{R}^{d}$ via weighted summation:
\begin{equation}
\mathbf{G}_{(i-1),j} = \sum_{m=1}^{n} \omega_{(i-1),j,m} \cdot \mathbf{V}_{(i-1),j,m},
\label{eq:global_feature}
\end{equation}
where $\mathbf{G}_{(i-1),j}$ captures contextual information across all image regions for the $(i-1)$-th task.

Subsequent denoising utilizes a Denoising Autoencoder (DAE) to refine $\mathbf{V}_{(i-1),j}$ into denoised features $\mathbf{V}'_{(i-1),j} \in \mathbb{R}^{n \times d}$. To further enhance the representation, the global feature $\mathbf{G}_{(i-1),j}$ and the denoised region features $\mathbf{V}'_{(i-1),j}$ are combined into the final enhanced features $\mathbf{V}''_{(i-1),j}$.

For optimization, GonF employs a unified loss function that simultaneously encourages accurate feature reconstruction and smooth attention distribution. Specifically, the loss is defined as the squared reconstruction error penalized by an entropy-based regularization term on the attention weights:
\begin{equation}
\begin{split}
\mathcal{L}_{\mathrm{GonF}} = 
\frac{1}{n} \sum_{m=1}^n \Big( 
    \|\mathbf{V}_{(i-1),j,m} - \mathbf{V}'_{(i-1),j,m}\|_2^2 \\
    - \theta_1 \, \omega_{(i-1),j,m} \log \omega_{(i-1),j,m} 
\Big)
\end{split},
\label{eq:gonf_loss_single}
\end{equation}
where $\theta_1$ is a hyperparameter (typically within $[0.1, 0.5]$) that balances the trade-off between denoising precision and attention distribution smoothness.

The enhanced visual features $\mathbf{V}''_{(i-1),j}$, together with textual features $\mathbf{Q}_{(i-1),j}$, are subsequently processed by the Multimodal Encoder, which leverages a Transformer backbone to fuse modalities into the final multimodal representation $\mathbf{F}_{(i-1),j} \in \mathbb{R}^{M \times d}$, where $M$ is the total number of fused tokens.

\subsection{Adaptive Memory Allocation}
\label{sec:3.2}

Conventional memory or prototype-based mechanisms often struggle to cope with the dynamic and heterogeneous nature of continual multimodal tasks. To overcome this limitation, we propose the Adaptive Memory Allocation (AMA) module, which enables the model to selectively leverage and update relevant prior knowledge. This design enhances generalization and robustness across a wide range of visual-question scenarios.

The input to AMA is a hidden state feature $\mathbf{H} \in \mathbb{R}^{d}$, which encodes the global context of a given sample. This feature is obtained from the multimodal encoder by fusing visual and textual information. To facilitate fine-grained prototype matching, $\mathbf{H}$ is projected into two modality-specific subspaces:
\begin{equation}
\mathbf{H}_v = \mathbf{W}_v \cdot \mathbf{H}, \quad \mathbf{H}_q = \mathbf{W}_q \cdot \mathbf{H},
\end{equation}
where $\mathbf{W}_v \in \mathbb{R}^{d_v \times d}$ and $\mathbf{W}_q \in \mathbb{R}^{d_q \times d}$ are learnable transformation matrices. Here, $\mathbf{H}_v \in \mathbb{R}^{d_v}$ and $\mathbf{H}_q \in \mathbb{R}^{d_q}$ represent the visual- and textual-specific features, respectively.

AMA maintains a memory pool consisting of two distinct prototype sets: visual prototypes $\{\mathbf{V}_m \in \mathbb{R}^{d_v} \mid m = 1, \dots, M\}$ and textual prototypes $\{\mathbf{Q}_n \in \mathbb{R}^{d_q} \mid n = 1, \dots, N\}$. Each incoming sample compares its modality-specific features to stored prototypes using cosine similarity:
\begin{equation}
\left\{
\begin{aligned}
\text{sim}(\mathbf{H}_v, \mathbf{V}_m) &= \frac{\mathbf{H}_v \cdot \mathbf{V}_m}{\|\mathbf{H}_v\| \cdot \|\mathbf{V}_m\|}, \\
\text{sim}(\mathbf{H}_q, \mathbf{Q}_n) &= \frac{\mathbf{H}_q \cdot \mathbf{Q}_n}{\|\mathbf{H}_q\| \cdot \|\mathbf{Q}_n\|},
\end{aligned}
\right.
\end{equation}
where $\cdot$ denotes dot product and $\|\cdot\|$ is the $L_2$ norm. The top-$k$ most similar visual and textual prototypes, denoted as $\{\mathbf{V}_{p_m}\}_{m=1}^k$ and $\{\mathbf{Q}_{p_n}\}_{n=1}^k$, are retrieved for further fusion.

To adaptively integrate retrieved prototypes with the current input, AMA employs a gating-based fusion mechanism. A dynamic weight vector $\mathbf{g} \in \mathbb{R}^{d_g}$ is computed as:
\begin{equation}
\mathbf{g} = \sigma(\mathbf{W}_g \cdot [\mathbf{H}; \mathbf{Q}_p; \mathbf{V}_p]),
\end{equation}
where $\mathbf{W}_g \in \mathbb{R}^{d_g \times (d + d_q + d_v)}$ is a learnable matrix, $\sigma$ is a non-linear activation (e.g., sigmoid), and $[\cdot;\cdot;\cdot]$ denotes vector concatenation. This gating mechanism allows the model to selectively emphasize useful modalities and ignore less relevant signals.

To ensure that knowledge is accumulated effectively across tasks, we further update the memory pool using a temporal interpolation strategy:
\begin{equation}
\mathbf{P}_i = \lambda \cdot \mathbf{P}_{i-1} + (1 - \lambda) \cdot \mathbf{H},
\label{eq:memory_update}
\end{equation}
where $\mathbf{P}_i$ denotes the updated memory state after processing task $i$, and $\lambda \in [0, 1]$ controls the balance between past memory $\mathbf{P}_{i-1}$ and current context $\mathbf{H}$. This incremental update rule ensures both long-term memory retention and adaptability to new information, making the memory pool more resilient to catastrophic forgetting.

The final fused feature $\mathbf{H}' \in \mathbb{R}^{d}$ is computed as:
\begin{equation}
\mathbf{H}' = \mathbf{H} + \boldsymbol{\alpha} \cdot \mathbf{Q}_p + \boldsymbol{\beta} \cdot \mathbf{V}_p,
\end{equation}
where $\boldsymbol{\alpha} \in \mathbb{R}^{d_q}$ and $\boldsymbol{\beta} \in \mathbb{R}^{d_v}$ are dynamically adjusted weights for the textual and visual prototypes. These weights are defined as:
\begin{equation}
\left\{
\begin{aligned}
\boldsymbol{\alpha} &= g_v \cdot \text{softmax}(\mathbf{W}_\alpha \cdot \mathbf{V}_p) \\
\boldsymbol{\beta} &= g_q \cdot \text{softmax}(\mathbf{W}_\beta \cdot \mathbf{Q}_p)
\end{aligned},
\right.
\end{equation}
where $g_v$ and $g_q$ are portions of $\mathbf{g}$ allocated to the visual and textual modalities, respectively. $\mathbf{W}_\alpha \in \mathbb{R}^{d_q \times d_v}$ and $\mathbf{W}_\beta \in \mathbb{R}^{d_v \times d_q}$ are learnable weight matrices, and $\text{softmax}$ ensures that the contributions of all prototypes are normalized to sum to 1.

The overall loss function for the AMA module is defined as:
\begin{align}
\mathcal{L}_{\text{AMA}} =\ 
    & -\sum_{m=1}^{k} \text{sim}(\mathbf{H}_v, \mathbf{V}_{p_m}) 
      - \sum_{n=1}^{k} \text{sim}(\mathbf{H}_q, \mathbf{Q}_{p_n}) \notag \\
    & +\ \theta_2 (g_q + g_v - 1)^2 
      + \theta_3 \|\mathbf{H}' - \mathbf{H}\|_2^2,
\end{align}
where $\theta_2$ and $\theta_3$ are hyperparameters used to balance the contributions of the regularization and update terms. In practice, $\theta_2$ and $\theta_3$ are typically selected from the range $[0.01, 1.0]$ based on validation performance.

\subsection{Multimodal Decoding for Answers}
\label{sec:3.3}

The multimodal decoding module generates task predictions based on fused visual and textual features. Given an input representation $\mathbf{H}' \in \mathbb{R}^{d}$, the model first applies two parallel linear projections:
\begin{equation}
\mathbf{E}_v = \mathbf{W}_v \mathbf{H}', \quad 
\mathbf{E}_q = \mathbf{W}_q \mathbf{H}',
\label{eq:projection}
\end{equation}
where $\mathbf{W}_v, \mathbf{W}_q \in \mathbb{R}^{d_e \times d}$ are learnable matrices, and $\mathbf{E}_v, \mathbf{E}_q \in \mathbb{R}^{d_e}$ are the corresponding visual and question embeddings.

\begin{table*}[!t]
\centering
\label{tab:all_results}
\definecolor{mypink}{RGB}{246,229,234}
\definecolor{mygray}{RGB}{224,224,224}
\definecolor{myy}{RGB}{255,250,230}

\small
\renewcommand{\arraystretch}{0.8}
\resizebox{\textwidth}{!}{%
\begin{tabular}{c|cccccccccc|cc}
\toprule
\multirow{2}{*}{\centering\textbf{Methods}} & \multicolumn{10}{c|}{\textbf{Various tasks in VQA v2}} & \multirow{2}{*}{\textbf{AP ($\uparrow$)}} & \multirow{2}{*}{\textbf{AF ($\downarrow$)}} \\ 
\cmidrule{2-11} 
& \textbf{Rec.} & \textbf{Loc.} & \textbf{Jud.} & \textbf{Com.} & \textbf{Cou.} & \textbf{Act.} & \textbf{Col.} & \textbf{Typ.} & \textbf{Sub.} & \textbf{Cau.} & & \\ \midrule

\multicolumn{1}{c}{\textbf{\textit{Standard Test}}} \\ \midrule
Vanilla  & 7.39 & 4.94 & 22.29 & 32.30 & 0.71 & 12.14 & 12.10 & 10.69 & 27.29 & 15.10 & 14.49 & 30.15 \\
EWC (PNAS'17) & 6.73 & 8.43 & 27.22 & 47.10 & 0.14 & 12.40 & 1.76 & 10.98 & 31.05 & 11.85 & 15.77 & 28.38 \\
  MAS (ECCV'18) & 30.81 & 8.07 & 25.50 & 4.00 & 31.90 & 32.39 & 26.24 & 24.75 & 19.85 & 2.75 & 20.56 & 21.97 \\
ER (MTLR'19) & 18.64 & 21.36 & 61.27 & 64.17 & 30.29 & 52.84 & 43.39 & 23.31 & 42.75 & 11.85 & 36.99 & 4.80 \\
DER (NeurIPS'20) & 14.55 & 13.83 & 62.88 & 65.16 & 30.96 & 51.19 & 40.51 & 19.04 & 42.87 & 12.55 & 35.35 & 6.58 \\
VQACL (CVPR'23) & 20.47 & 28.02 & 62.55 & 68.61 & 29.35 & 50.66 & 44.45 & 26.36 & 44.65 & 12.60 & 38.77 & 2.90 \\
QUAD (arXiv'25) & 20.51 & 28.04 &\cellcolor{mygray}\underline{62.62} & {68.65} & 29.39 & 50.67 & 44.48 & 26.38 & 44.72 & 12.64 & 39.25 & 3.91 \\
ProtoGroup (ICASSP'25) &\cellcolor{mygray}\underline{20.66} &\cellcolor{mygray}\underline{28.77} & 62.55 &\cellcolor{mypink}\textbf{69.21} &\cellcolor{mygray}\underline{29.44} &\cellcolor{mygray}\underline{50.79} &\cellcolor{mygray}\underline{44.66} &\cellcolor{mygray}\underline{26.44} &\cellcolor{mygray}\underline{44.86} &\cellcolor{mygray}\underline{12.70} &\cellcolor{mygray}\underline{39.81} &\cellcolor{mygray}\underline{2.87} \\ 
\noalign{\vskip 0.8ex \hrule height 0.4pt \vskip 0.5ex}  
MacVQA (Ours) &\cellcolor{mypink}\bftab 26.19 &\cellcolor{mypink}\bftab 28.96 &\cellcolor{mypink}\bftab 66.64 &\cellcolor{mygray}\underline{66.22} &\cellcolor{mypink} \bftab 32.30 &\cellcolor{mypink} \bftab 58.13 &\cellcolor{mypink}\bftab 56.95 &\cellcolor{mypink}\bftab 35.90 &\cellcolor{mypink}\bftab 50.78 &\cellcolor{mypink}\bftab 15.95 &\cellcolor{mypink}\bftab 43.38 &\cellcolor{mypink}\bftab 2.32 \\ \midrule

\multicolumn{1}{c}{\textbf{\textit{Novel Comp. Test}}} \\ \midrule
Vanilla  & 7.23 & 4.55 & 21.56 & 31.98 & 0.66 & 12.08 & 11.88 & 8.99 & 27.12 & 14.70 & 11.79 & 27.16 \\
EWC (PNAS'17) & 6.44 & 8.14 & 26.93 & 46.81 & 0.05 & 12.02 & 1.67 & 10.60 & 30.76 & 11.56 & 12.83 & 28.16 \\
  MAS (ECCV'18) & 31.06 & 8.32 & 25.75 & 4.25 & 32.15 & 32.64 & 26.49 & 25.00 & 20.10 & 3.00 & 23.90 & 6.24 \\
ER (MTLR'19) & 18.32 & 21.04 & 60.79 & 64.01 & 29.97 & 52.52 & 43.07 & 22.99 & 42.43 & 11.53 & 33.78 & 5.76 \\
DER (NeurIPS'20) & 14.17 & 13.45 & 62.50 & 64.78 & 30.58 & 50.81 & 39.94 & 18.85 & 42.49 & 12.17 & 31.52 & 8.59 \\
VQACL (CVPR'23) & 20.13 & 27.68 & 62.38 & 68.10 & 29.01 & 50.32 & 44.11 & 26.02 & 44.31 & 12.26 & 35.40 & 4.90 \\
QUAD (arXiv'25) &\cellcolor{mygray}\underline{20.59} &\cellcolor{mygray}\underline{28.08} &\cellcolor{mygray}\underline{63.01} &\cellcolor{mygray}\underline{68.73} &\cellcolor{mygray}\underline{29.47} &\cellcolor{mygray}\underline{50.75} &\cellcolor{mygray}\underline{44.56} &\cellcolor{mygray}\underline{26.46} &\cellcolor{mygray}\underline{44.80} &\cellcolor{mygray}\underline{12.72} &\cellcolor{mygray}\underline{40.00} &\cellcolor{mygray}\underline{3.81} \\
ProtoGroup (ICASSP'25) & 20.33 & 27.89 & 62.22 & 68.64 & 29.23 & 50.45 & 44.27 & 26.14 & 44.42 & 12.34 & 36.81 & 4.09 \\ 
\noalign{\vskip 0.8ex \hrule height 0.4pt \vskip 0.5ex}  
MacVQA (Ours) &\cellcolor{mypink}\bftab 24.35 &\cellcolor{mypink}\bftab 28.09 &\cellcolor{mypink}\bftab 67.05 &\cellcolor{mypink}\bftab 69.65 &\cellcolor{mypink}\bftab 30.06 &\cellcolor{mypink}\bftab 48.41 &\cellcolor{mypink}\bftab 54.05 &\cellcolor{mypink}\bftab 29.43 &\cellcolor{mypink}\bftab 47.21 &\cellcolor{mypink}\bftab 17.00 &\cellcolor{mypink}\bftab 42.53 &\cellcolor{mypink}\bftab 3.60 \\ \bottomrule
\end{tabular}
}
\caption{Comparison of various tasks in VQA v2 across Standard Test and Novel Compositional Test.}
\label{tab:vqa_comparison}
\end{table*}

\begin{table*}[!t]
\centering
\label{tab:ablation_results}
\definecolor{mypink}{RGB}{246,229,234}
\small
\renewcommand{\arraystretch}{0.85}
\begin{tabular}{c|c|c|c|c|c|c|c|c|c|c|c|c|c}
\toprule
\multicolumn{2}{c|}{\textbf{Method}} & \multicolumn{10}{c|}{\textbf{Various task in VQA v2}} & \multirow{2}{*}{\textbf{AP ($\uparrow$)}} & \multirow{2}{*}{\textbf{AF ($\downarrow$)}} \\
\cmidrule{1-2} \cmidrule{3-12} 
\textbf{GonF} & \textbf{AMA} & \textbf{Rec.} & \textbf{Loc.} & \textbf{Jud.} & \textbf{Com.} & \textbf{Cou.} & \textbf{Act.} & \textbf{Col.} & \textbf{Typ.} & \textbf{Sub.} & \textbf{Cau.} & & \\
\midrule
× & × & 20.47 & 28.02 & 62.55 & 68.61 & 29.35 & 50.66 & 44.45 & 26.36 & 44.65 & 12.60 & 38.77 & 2.90 \\
\checkmark & × & 23.65 &\cellcolor{mypink}\textbf{36.08} & 66.09 & 65.90 & 32.24 & 47.68 & 55.37 & 32.09 & 44.76 & 14.30 & 41.75 & 2.14 \\
× & \checkmark & 22.54 & 35.05 & 65.79 & 65.90 & 32.15 & 45.30 & 55.97 & 31.96 & 44.05 & 13.78 & 40.97 & 2.34 \\ 
\checkmark & \checkmark &\cellcolor{mypink}\textbf{26.19} & 28.96 &\cellcolor{mypink}\textbf{66.64} &\cellcolor{mypink}\textbf{66.22} &\cellcolor{mypink}\textbf{32.30} &\cellcolor{mypink}\textbf{58.13} &\cellcolor{mypink}\textbf{56.95} &\cellcolor{mypink}\textbf{35.90} &\cellcolor{mypink}\textbf{50.78} &\cellcolor{mypink}\textbf{15.95} &\cellcolor{mypink}\textbf{43.38\tiny{(+3.57)}} &\cellcolor{mypink} \textbf{2.32\tiny{(-0.58)}} \\
\bottomrule
\end{tabular}
\caption{Ablation study of GonF and AMA modules on VQA performance.}
\label{tab:ablation_results}
\end{table*}

These embeddings are concatenated to form a unified decoder input:
\begin{equation}
\mathbf{E} = [\mathbf{E}_v; \mathbf{E}_q],
\label{eq:concat}
\end{equation}
where $\mathbf{E} \in \mathbb{R}^{2d_e}$ combines modality-specific cues into a single feature for downstream VQA prediction.

The decoder receives the fused multimodal embeddings $\mathbf{E}$ and generates task-specific outputs, such as answers to visual questions. It integrates visual and textual information through the cross-attention mechanism, expressed as:
\begin{equation}
\text{Attention}(\mathbf{Q}, \mathbf{K}, \mathbf{V}) = \text{softmax}\left(\frac{\mathbf{Q}\mathbf{K}^\top}{\sqrt{d_k}}\right)\mathbf{V},
\end{equation}
where $\mathbf{Q}$, $\mathbf{K}$, and $\mathbf{V}$ represent the query, key, and value matrices derived from the multimodal embeddings $\mathbf{E}$, and $d_k$ is the dimensionality of the key vectors.

During training, the decoder minimizes the negative log-likelihood of the ground truth answer $A_{\text{true}}$:
\begin{equation}
\mathcal{L}_{\text{decoder}} = -\sum_{t=1}^{T} \log P(A_{\text{true}, t} \mid \mathbf{Q}, \mathbf{I}, A_{\text{true}, <t}),
\end{equation}
where $T$ is the length of the answer sequence, and $A_{\text{true}, <t}$ represents the partial sequence of the ground truth answer up to time step $\textit{t}-1$.

The total loss of the framework is defined as:
\begin{equation}
\mathcal{L}_{\text{total}} = \phi_1 \mathcal{L}_{\text{GonF}} + \phi_2 \mathcal{L}_{\text{AMA}} + \phi_3 \mathcal{L}_{\text{decoder}},
\end{equation}
where $\phi_1$, $\phi_2$, and $\phi_3$ are hyperparameters balancing the contributions of the three modules, and satisfy $\phi_1, \phi_2, \phi_3 \geq 0$ and $\phi_1 + \phi_2 + \phi_3 = 1$.

In summary, the Multimodal Decoding module integrates visual and textual information through cross-attention to generate accurate answers. Optimizing and balancing the total loss ensures robust multimodal understanding and enhances overall framework performance.

\section{Experiments}
To comprehensively evaluate our proposed MacVQA framework, we conduct extensive experiments addressing six core research questions:

\begin{itemize}
    \item \textbf{RQ1:} How does MacVQA's performance compare with baselines under both standard and novel composition training paradigms?
    \item \textbf{RQ2:} What contribution does each core component make to MacVQA's overall effectiveness?
    \item \textbf{RQ3:} How do different prototype selection strategies impact model accuracy and stability?
    \item \textbf{RQ4:} How does memory buffer capacity influence knowledge retention and forgetting in MacVQA?
    \item \textbf{RQ5:} How sensitive is MacVQA to key hyperparameter configurations?
    \item \textbf{RQ6:} How robustly does MacVQA generalize to real-world datasets with dynamically evolving tasks?
\end{itemize}

\subsection{Experimental Settings}
\label{sec:4.1}

\noindent\textbf{Datasets}~~~We evaluate on the VQA v2 benchmark, which includes over 200k COCO images \cite{34} and 1.1M annotated QA pairs. Following VQACL, the dataset is partitioned into 10 tasks by question type to simulate real-world sequential learning (recognition, location, judge, commonsense, count, action, color, type, subcategory, and causal.).

\begin{table}[!t] 
\centering
\label{tab:prototype_strategies}
\renewcommand{\arraystretch}{0.85}
\definecolor{mypink}{RGB}{246,229,234}
\renewcommand{\arraystretch}{0.9}
\small
\begin{tabular}{c|c|c|c|c|c}
\toprule
\multirow{2}{*}{\textbf{Method}} & \multirow{2}{*}{\textbf{Mechanism}} & \multicolumn{2}{c|}{\textbf{standard}} & \multicolumn{2}{c}{\textbf{novel}} \\
\cmidrule(r){3-4} \cmidrule(l){5-6}
 & & \textbf{AP} & \textbf{AF} & \textbf{AP} & \textbf{AF} \\
\midrule
\multirow{2}{*}{ProtoGroup} & Random & 37.22 & 3.89 & 35.78 & 4.54 \\
 & Max-Similarity &\cellcolor{mypink}\bftab 39.81 &\cellcolor{mypink}\bftab 2.87 &\cellcolor{mypink}\bftab 36.81 &\cellcolor{mypink}\bftab 4.09 \\
 \midrule
\multirow{2}{*}{MacVQA} & Random & 40.29 & 3.46 & 41.25 & 3.76 \\
 & Max-Similarity &\cellcolor{mypink}\bftab 43.38 &\cellcolor{mypink}\bftab 2.32 &\cellcolor{mypink}\bftab 42.53 &\cellcolor{mypink}\bftab 3.60 \\
\bottomrule
\end{tabular}
\caption{Performance comparison of prototype selection strategies (Random vs. Max-Similarity) under standard and novel composition scenarios.}
\label{tab:prototype_strategies}
\end{table}

\begin{figure}[!t]
\centering
\includegraphics[width=1.0\columnwidth]{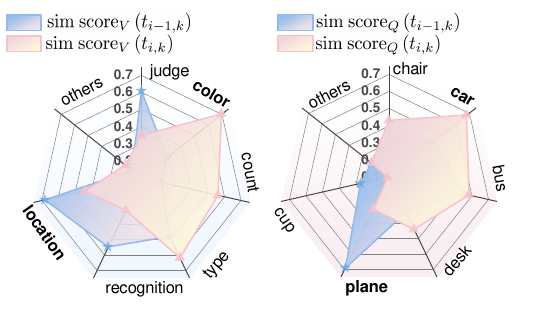} 
\caption{Similarity score radar plots for two examples (the \textit{k}-th data from task \textit{i}-1 and task \textit{i}). Left: question memory; right: visual memory. Top-3 prototypes are selected by similarity.}
\label{figure:3}
\end{figure}

\begin{figure}[!t]
\centering
\includegraphics[width=1.0\columnwidth]{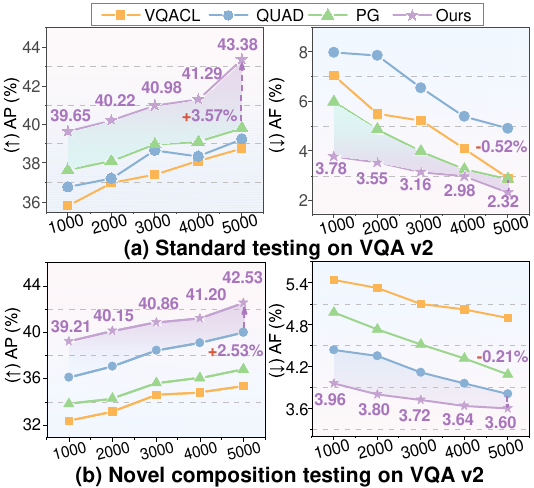} 
\caption{Memory size sensitivity analysis for VQA v2 under standard and novel composition testing paradigms}
\label{figure:4}
\end{figure}
\noindent\textbf{Evaluation Metrics}~~~We report Final Average Performance (AP) and Average Forgetting (AF)~\cite{32}, which respectively reflect the overall accuracy after continual learning and the degradation on previously learned tasks.

\noindent\textbf{Baselines}~~~Methodologies are categorized by core mechanisms: (1) Vanilla (sequential fine-tuning without continual learning mechanisms, serving as lower-bound benchmark); (2) Regularization-based methods: EWC \cite{42},   MAS \cite{11}; (3) Replay-based methods: ER \cite{13}, DER \cite{14}, , VQACL \cite{15}, QUAD \cite{16}, ProtoGroup \cite{17}. 

\noindent\textbf{Implementation Details}~~~All methods use the same pretrained VL-T5 backbone~\cite{20}. Evaluation follows two paradigms: standard testing on seen skill–concept combinations, and novel composition testing on held-out combinations. A dynamically optimized memory buffer is employed. We use Adam~\cite{53} with a 3e-5 learning rate, gradient clipping of 5, and a 0.1 warmup ratio.

\subsection{Main Results (RQ1)}
\label{sec:4.2}
As shown in Table~\ref{tab:vqa_comparison}, MacVQA achieves an AP of 43.38\% and an AF of 2.32\% on the Standard Test, outperforming all baseline methods. On the Novel Composition Test, it maintains strong results with 42.53\% AP and 3.60\% AF. In standard settings, MacVQA achieves the highest accuracy on tasks such as \textit{judgment} (66.64\%), \textit{action} (58.13\%), and \textit{color} (56.95\%), reflecting its strength in both reasoning and perception. Under the more challenging novel setting, it delivers top performance in \textit{commonsense} reasoning with 67.05\% accuracy, and shows clear advantages in \textit{subcategory} classification and \textit{causal} reasoning with 47.21\% and 17.00\%, respectively. These results confirm MacVQA’s ability to retain prior knowledge, adapt to new tasks, and generalize effectively across all ten continual VQA tasks. This highlights the effective coordination between visual filtering and memory adaptation components.

\subsection{Effects of Modality (RQ2)}
\label{sec:4.3}

Table~\ref{tab:ablation_results} reports the ablation study on the Global Noise Filtering (GonF) and Adaptive Memory Allocation (AMA) modules. Introducing either module individually improves performance over the baseline, with GonF reducing AF from 2.90\% to 2.14\% and AMA increasing AP from 38.77\% to 40.97\%. When combined, the full MacVQA model achieves the best results with 43.38\% AP and 2.32\% AF. GonF enhances perception tasks such as recognition and location, while AMA supports memory-sensitive reasoning like color and typicality. GonF’s early filtering curbs error propagation, and AMA promotes long-term retention by prioritizing salient prototypes. Together, they form a complementary mechanism that balances feature robustness and continual learning.

\subsection{Prototype Selection Strategies (RQ3)}
\label{sec:4.4}

As shown in Table~\ref{tab:prototype_strategies}, prototype selection plays a crucial role in continual VQA. Compared to random sampling, the Max-Similarity strategy consistently yields better performance across both standard and novel settings. On the Standard Test, it achieves 43.38\% AP and 2.32\% AF, outperforming the random variant by notable margins. Figure~\ref{figure:3} shows that selecting top-3 prototypes by similarity enables retrieval of semantically aligned exemplars, improving rehearsal relevance and model stability. These results confirm that informed prototype selection is key to effective knowledge retention and generalization.

\begin{figure}[!t]
\centering
\includegraphics[width=1.0\columnwidth]{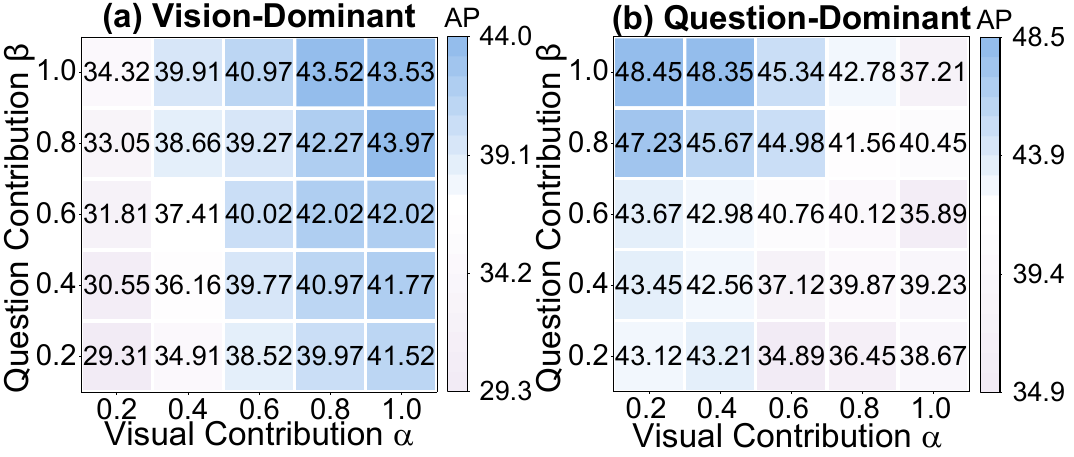} 
\caption{Task-Specific Hyperparameter Sensitivity of MacVQA.}
\label{figure:5}
\end{figure}

\subsection{Memory Capacity Sensitivity Analysis (RQ4)}
\label{sec:4.5}

Figure~\ref{figure:4} illustrates the impact of memory size on model performance under both standard and novel composition settings. While all methods benefit from increased buffer capacity, the degree and stability of improvement vary. VQACL shows rapid AP gains at smaller sizes but suffers from fluctuating AF, while QUAD consistently exhibits high AF, indicating inefficient memory use. ProtoGroup yields modest gains. In contrast, MacVQA maintains the highest AP and lowest AF across all memory scales. At the largest buffer size (5000), it achieves 43.38\% AP and 2.32\% AF on the standard test, and 42.53\% AP with 3.60\% AF on the novel test, outperforming the best baseline by +3.57\% AP and –0.52\% AF. These findings underscore MacVQA’s efficient memory use and its robustness across varying capacities.

\begin{figure}[t]
\centering
\includegraphics[width=1.0\linewidth]{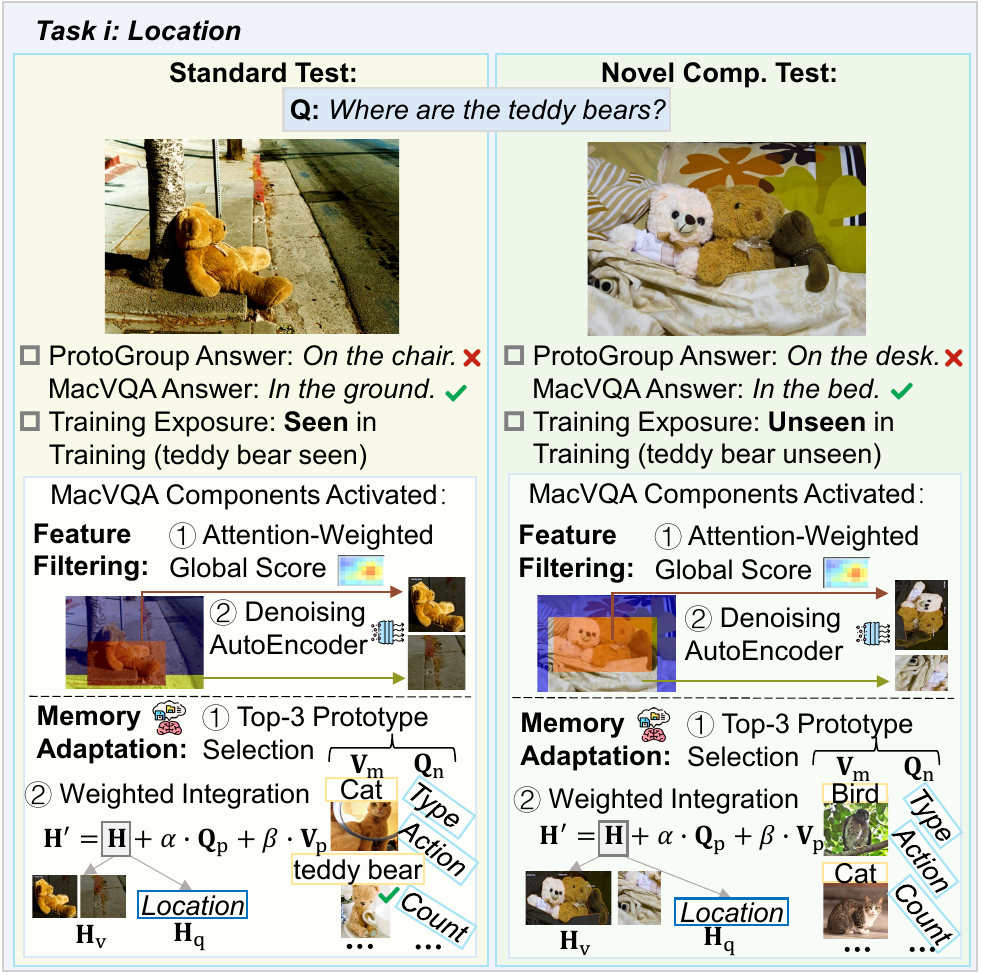} 
\caption{MacVQA exemplars in the \textit{Location} task under standard and novel test settings.}
\label{figure:6}
\end{figure}

\subsection{Hyperparameter Sensitivity Analysis (RQ5)}
\label{sec:4.6}

Figure~\ref{figure:5} presents the sensitivity analysis of MacVQA under different combinations of visual contribution $\alpha$ and question contribution $\beta$, showing clear task-specific preferences. For visual-dominant tasks such as recognition or location (Figure~\ref{figure:5}(a)), performance peaks at $\alpha=1.0$, $\beta=0.4$ with an AP of 43.53\%. In contrast, question-dominant tasks such as commonsense or judgment (Figure~\ref{figure:5}(b)) favor higher textual weights, with a best performance of 48.5\% at $\alpha=0.4$, $\beta=1.0$. This highlights MacVQA’s flexibility in adapting to diverse reasoning needs through multimodal fusion. These findings support the use of modality-aware weighting and validate the design of MacVQA’s dynamic prototype integration.

\subsection{Real-world Robustness Analysis (RQ6)}
\label{sec:4.7}

Figure~\ref{figure:6} presents qualitative exemplars that illustrate MacVQA’s robustness across both standard and novel testing scenarios. In the \textit{location} task, the model correctly answers “\textit{in the chair}” for a standard test image where the object type (teddy bear) was seen during training. It also succeeds on a novel case involving an unseen combination—answering “\textit{in the bed}” when the teddy bear type had not appeared in training for this reasoning context. In contrast, baseline methods fail in both settings. These examples highlight MacVQA’s ability to generalize to unseen object-reasoning pairs while maintaining accuracy on familiar inputs. This stems from its feature filtering mechanism, which reduces visual noise, and its adaptive memory strategy, which retrieves semantically aligned prototypes. Together, these components enable robust performance under distribution shifts in continual learning.

\section{Conclusion}
We propose MacVQA to address the challenges of balancing knowledge retention, adaptation efficiency, and robust reasoning in multimodal continual learning. By introducing Global Noise Filtering for robust feature representation and Adaptive Memory Allocation for efficient and effective memory management, MacVQA achieves a balanced fusion of visual and textual information while mitigating the impact of noise. Experiments on ten continual VQA tasks from the VQA v2 dataset demonstrate significant improvements in both accuracy and resistance to forgetting, validating the effectiveness of MacVQA. Future work will extend this framework to broader modalities and explore dynamic memory strategies for large-scale scenarios.

\section{Acknowledgments}
This work was supported in part by the National Natural Science Foundation of China (No. 62207011, 62407013, 62377009, 62101179), the Natural Science Foundation of Hubei Province of China (No. 2025AFB653), the Natural Science Foundation of Shandong Province of China (No. ZR2024QF257), the Science and Technology Support Plan for Youth Innovation of Colleges and Universities of Shandong Province of China (No. 2023KJ370), the Open Fund of Hubei Key Laboratory of Big Data Intelligent Analysis and Application, Hubei University (No. 2024BDIAA05), and the Open Fund of Key Laboratory of Intelligent Sensing System and Security of Hubei University, Ministry of Education (No. KLISSS202410).

\bibliography{aaai2026}

\end{document}